\documentclass[10pt,twocolumn,letterpaper]{article}

\usepackage{iccv}              %

\definecolor{iccvblue}{rgb}{0.21,0.49,0.74}
\usepackage[pagebackref,breaklinks,colorlinks,allcolors=iccvblue]{hyperref}
\usepackage{tabularray}
\usepackage{graphicx} %
\usepackage{tabularx} %
\usepackage{array}
\usepackage{booktabs}
\usepackage{arydshln}
\usepackage{multirow}
\usepackage{makecell}
\usepackage{tikz}
\usepackage{pifont}
\usepackage[accsupp]{axessibility}  %

\def\paperID{14036} %
\def\confName{ICCV}
\def\confYear{2025}

\title{YOLOE: Real-Time Seeing Anything}

\author{Ao Wang$^{1}$\thanks{Equal contributions. $\dagger$ Corresponding author.} \quad Lihao Liu$^{1}$\footnotemark[1] \quad Hui Chen$^{2}$ \quad Zijia Lin$^{1}$ \quad Jungong Han$^{3}$ \quad Guiguang Ding$^{1,\dagger}$ \\
		$^1$School of Software, Tsinghua University \quad $^2$BNRist, Tsinghua University \\$^3$Department of Automation, Tsinghua University\\
        }

\begin{document}
\maketitle
\begin{abstract}
Object detection and segmentation are widely employed in computer vision applications, yet conventional models like YOLO series, while efficient and accurate, are limited by predefined categories, hindering adaptability in open scenarios. Recent open-set methods leverage text prompts, visual cues, or prompt-free paradigm to overcome this, but often compromise between performance and efficiency due to high computational demands or deployment complexity. In this work, we introduce YOLOE, which integrates detection and segmentation across diverse open prompt mechanisms within a single highly efficient model, achieving real-time seeing anything. For text prompts, we propose Re-parameterizable Region-Text Alignment (RepRTA) strategy. It refines pretrained textual embeddings via a re-parameterizable lightweight auxiliary network and enhances visual-textual alignment with zero inference and transferring overhead. For visual prompts, we present Semantic-Activated Visual Prompt Encoder (SAVPE). It employs decoupled semantic and activation branches to bring improved visual embedding and accuracy with minimal complexity. For prompt-free scenario, we introduce Lazy Region-Prompt Contrast (LRPC) strategy. It utilizes a built-in large vocabulary and specialized embedding to identify all objects, avoiding costly language model dependency. Extensive experiments show YOLOE's exceptional zero-shot performance and transferability with high inference efficiency and low training cost. Notably, on LVIS, with 3$\times$ less training cost and 1.4$\times$ inference speedup, YOLOE-v8-S surpasses YOLO-Worldv2-S by 3.5 AP. When transferring to COCO, YOLOE-v8-L achieves 0.6 AP$^b$ and 0.4 AP$^m$ gains over closed-set YOLOv8-L with nearly 4$\times$ less training time. Code and models are available at \href{https://github.com/THU-MIG/yoloe}{here}.
\begin{tikzpicture}[remember picture,overlay,shift={(current page.north west)}]
    \node[anchor=north west,xshift=5.8cm,yshift=-3.48cm]{\includegraphics[height=0.04\textwidth]{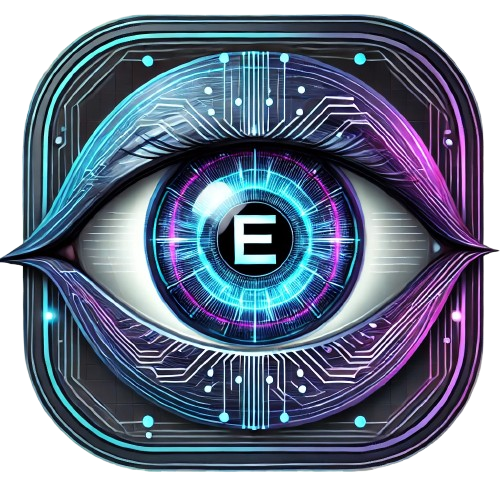}};
\end{tikzpicture}
\end{abstract}

\section{Introduction}
Object detection and segmentation are foundational tasks in computer vision~\cite{ren2015faster,he2017mask}, with widespread applications spanning autonomous driving~\cite{bogdoll2022anomaly}, medical analyses~\cite{sobek2024medyolo}, and robotics~\cite{dos2019mobile}, \etc. Traditional approaches like YOLO series~\cite{redmon2016you,bochkovskiy2020yolov4,jocherultralytics,bolya2019yolact}, have leveraged convolutional neural networks to achieve real-time remarkable performance. However, their dependence on predefined object categories constrains flexibility in practical open scenarios. Such scenarios increasingly demand models capable of detecting and segmenting arbitrary objects guided by diverse prompt mechanisms, such as texts, visual cues, or without prompt.

\begin{figure}[t]
    \centering
    \includegraphics[width=0.95\columnwidth]{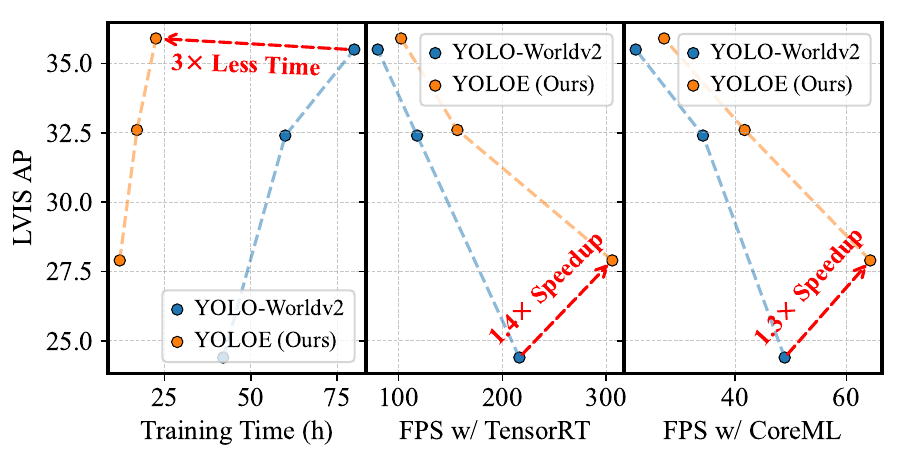}
    \caption{Comparison of performance, training cost, and inference efficiency between YOLOE (Ours) and advanced YOLO-Worldv2 in terms of open text prompts. LVIS AP is evaluated on \texttt{minival} set and FPS w/ TensorRT and w/ CoreML is measured on T4 GPU and iPhone 12, respectively. The results highlight our superiority.}
    \label{fig:comparison}
    \vspace{-16pt}
\end{figure}

Given this, recent efforts have shifted towards enabling models to generalize for open prompts~\cite{ren2024dinox,jiang2024trex2,zou2023seem,cheng2024yoloworld}. They target single prompt type, \eg, GLIP~\cite{li2022glip}, or multiple prompt types in a unified way, \eg, DINO-X~\cite{ren2024dinox}. Specifically, with region-level vision-language pretraining~\cite{li2022glip,liu2024gdino,yao2022detclip}, text prompts are usually processed by text encoder to serve as contrastive objectives for region features~\cite{ren2024dinox,jiang2024trex2}, achieving recognition for arbitrary categories, \eg, YOLO-World~\cite{cheng2024yoloworld}. For visual prompts, they are often encoded as class embeddings tied to specified regions for identifying similar objects, by the interaction with image features or language-aligned visual encoder~\cite{jiang2023t,li2024dinov,ren2024dinox,cheng2024yoloworld}, \eg, T-Rex2~\cite{jiang2024trex2}. In prompt-free scenario, existing methods typically integrate language models, finding all objects and generating the corresponding category names conditioned on region features sequentially~\cite{ren2024dinox,wu2024grit}, \eg, GenerateU~\cite{lin2024generateu}.

Despite notable advancements, a single model that supports diverse open prompts for arbitrary objects with high efficiency and accuracy is still lacking. For example, DINO-X~\cite{ren2024dinox} features a unified architecture, which, however, incurs resource-intensive training and inference overhead. Additionally, individual designs for different prompts in separate works exhibit suboptimal trade-offs between performance and efficiency, making it difficult to directly combine them into one model. For example, text-prompted approaches often incur substantial computational overhead when incorporating large vocabularies, due to complexity of cross-modality fusion~\cite{cheng2024yoloworld,liu2024gdino,li2022glip,ren2024dinox}. Visual-prompted methods usually compromise deployability on edge devices owing to the transformer-heavy design or reliance on additional visual encoder~\cite{jiang2024trex2,li2024dinov,zang2022ovdetr}. Prompt-free ways, meanwhile, depend on large language models, introducing considerable memory and latency costs~\cite{lin2024generateu,ren2024dinox}.

In light of these, in this paper, we introduce YOLOE(ye), a highly \textbf{efficient}, \textbf{unified}, and \textbf{open} object detection and segmentation model, like human eye, under different prompt mechanisms, like texts, visual inputs, and prompt-free paradigm. We begin with YOLO models with widely proven efficacy. For text prompts, we propose a Re-parameterizable Region-Text Alignment (RepRTA) strategy, which employs a lightweight auxiliary network to improve pretrained textual embeddings for better visual-semantic alignment. During training, pre-cached textual embeddings require only the auxiliary network to process text prompts, incurring low additional cost compared with closed-set training. At inference and transferring, auxiliary network is seamlessly re-parameterized into the classification head, yielding an architecture identical to YOLOs with zero overhead. For visual prompts, we design a Semantic-Activated Visual Prompt Encoder (SAVPE). By formalizing regions of interest as masks, SAVPE fuses them with multi-scale features from PAN to produce grouped prompt-aware weights in low dimension in an activation branch and extract prompt-agnostic semantic features in a semantic branch. Prompt embeddings are derived through aggregation of them, resulting in favorable performance with minimal complexity. For prompt-free scenario, we introduce Lazy Region-Prompt Contrast (LRPC) strategy. Without relying on costly language models, LRPC leverages a specialized prompt embedding to find all objects and a built-in large vocabulary for category retrieval. By matching only anchor points with identified objects against the vocabulary, LRPC ensures high performance with low overhead.

Thanks to them, YOLOE excels in detection and segmentation across diverse open prompt mechanisms within one model, enjoying high inference efficiency and low training cost. Notably, as shown in \cref{fig:comparison}, under 3$\times$ less training cost, YOLOE-v8-S significantly outperforms YOLO-Worldv2-S~\cite{cheng2024yoloworld} by 3.5 AP on LVIS~\cite{gupta2019lvis}, with 1.4$\times$ and 1.3$\times$ inference speedups on T4 and iPhone 12, respectively. In visual-prompted and prompt-free settings, YOLOE-v8-L outperforms T-Rex2 by 3.3 AP$_r$ and GenerateU by 0.4 AP with 2$\times$ less training data and 6.3$\times$ fewer parameters, respectively. For transferring to COCO~\cite{lin2014mscoco}, YOLOE-v8-M / L outperforms YOLOv8-M / L by 0.4 / 0.6 AP$^b$ and 0.4 / 0.4 AP$^m$ with nearly 4$\times$ less training time. We hope that YOLOE can establish a strong baseline and inspire further advancements in real-time open prompt-driven vision tasks.

\section{Related Work}
\textbf{Traditional detection and segmentation.} Traditional approaches for object detection and segmentation primarily operate under closed-set paradigms. Early two-stage frameworks~\cite{girshick2015fast,ren2015faster,he2017mask,cai2018cascade}, exemplified by Faster R-CNN~\cite{ren2015faster}, introduce region proposal networks (RPNs) followed by region-of-interest (ROI) classification and regression. Meanwhile, single-stage detectors~\cite{liu2016ssd,lin2017focal,tian2019fcos,zhang2020atss,feng2021tood} prioritizes speed through grid-based predictions within a single network. The YOLO series~\cite{redmon2016you,bochkovskiy2020yolov4,li2022yolov6,wang2023yolov7,jocherultralytics,wang2025yolov10} plays a significant role in this paradigm and are widely used in real world. Moreover, DETR~\cite{li2022dndetr} and its variants~\cite{zhu2020deformabledetr,li2022dndetr,zhang2022dino} mark a major shift by removing heuristic-driven components with transformer-based architectures. To achieve finer-grained results, existing instance segmentation methods predict pixel-level masks rather than bounding box coordinates~\cite{he2017mask}. For this, YOLACT~\cite{bolya2019yolact} facilitates real-time instance segmentation through integration of prototype masks and mask coefficients. Based on DINO~\cite{zhang2022dino}, MaskDINO~\cite{li2023maskdino} utilizes query embeddings and a high-resolution pixel embedding map to produce binary masks.

\begin{figure*}[t]
    \centering
    \includegraphics[width=0.9\textwidth]{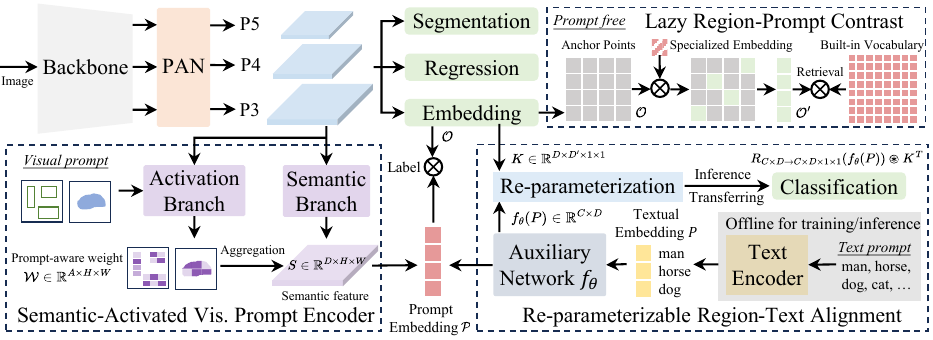}
    \caption{The overview of YOLOE, which supports detection and segmentation for diverse open prompt mechanisms. For text prompts, We design a re-parameterizable region-text alignment strategy to improve performance with zero inference and transferring overhead. For visual prompts, SAVPE is employed to encode visual cues with enhanced prompt embedding under minimal cost. For prompt-free setting, we introduce lazy region-prompt contrast strategy to provide category names for all identified objects efficiently by retrieval.}
    \label{fig:pipeline}
    \vspace{-10pt}
\end{figure*}

\textbf{Text-prompted detection and segmentation.}
Recent advancements in open-vocabulary object detection~\cite{zareian2021ovrcnn, gu2021vild, zhou2022detic, zhong2022regionclip, kuo2022fvlm,wang2024ovlw,zhao2024real,yao2022detclip} have focused on detecting novel categories by aligning visual features with textual embeddings. Specifically, GLIP~\cite{li2022glip} unifies object detection and phrase grounding through grounded pre-training on large-scale image-text pairs, demonstrating robust zero-shot performance. 
Grounding DINO~\cite{liu2024gdino} enhances this by integrating cross-modality fusion into DINO, improving alignment between text prompts and visual representations. YOLO-World~\cite{cheng2024yoloworld} further shows the potential of pretraining small detectors with open recognition capabilities based on the YOLO architecture. YOLO-UniOW~\cite{liu2024yolouniow} builds upon YOLO-World by leveraging the adaptive decision-learning strategy. Similarly, several open-vocabulary instance segmentation models~\cite{li2022lseg, ghiasi2022openseg, huynh2022open, xu2022groupvit, rao2022denseclip} learn rich visual-semantic knowledge from advanced foundation models to perform segmentation on novel object categories. For example, X-Decoder~\cite{zou2023xdecoder} and OpenSeeD~\cite{zhang2023openseed} explore both the open-vocabulary detection and segmentation tasks. APE~\cite{shen2024ape} introduces universal visual perception model that aligns and prompts all objects using various text prompts.

\textbf{Visual-prompted detection and segmentation.} While text prompts offer a generic description, certain objects can be challenging to describe with language alone, such as those requiring specialized domain knowledge. In such cases, visual prompts can guide detection and segmentation more flexibly and specifically, complementing text prompts~\cite{jiang2023t,jiang2024trex2,kirillov2023sam}. OV-DETR~\cite{zang2022ovdetr} and OWL-ViT~\cite{minderer2022owlvit} leverage CLIP encoders to process text and image prompts. 
DINOv~\cite{li2024dinov} explores visual prompts as in-context examples for generic and referring vision tasks. T-Rex2~\cite{jiang2024trex2} integrates visual and text prompts by region-level contrastive alignment. For segmentation, 
SEEM~\cite{zou2023seem} explores segmenting objects with various prompt types. Semantic-SAM~\cite{li2024semanticsam} excels in semantic comprehension and granularity detection, handling both panoptic and part segmentation.

\textbf{Prompt-free detection and segmentation.} Existing approaches still depend on explicit prompts during inference for open-set detection and segmentation. To address this limitation, several works~\cite{wu2024grit,long2023capdet,yao2024detclipv3,lin2024generateu,ren2024dinox} explore integrating with generative language models to produce object descriptions for all found objects. For instance, GRiT~\cite{wu2024grit} employs a text decoder for both dense captioning and object detection tasks. DetCLIPv3~\cite{yao2024detclipv3} trains an object captioner on large-scale data, enabling model to generate rich label information. GenerateU~\cite{lin2024generateu} leverages the language model to generate object names in a free-form way. 

\textbf{Closing remarks.} To the best of our knowledge, aside from DINO-X~\cite{ren2024dinox}, few efforts have achieved object detection and segmentation across various open prompt mechanisms within a single architecture. However, DINO-X entails extensive training cost and notable inference overhead, severely constraining the practicality for real-world edge deployments. In contrast, our YOLOE aims to deliver an efficient and unified model that enjoys real-time performance and efficiency with easy deployability.

\section{Methodology}
In this section, we detail designs of YOLOE. Building upon YOLOs (\cref{sec:architecture}), YOLOE supports text prompts through RepRTA (\cref{sec:reprta}), visual prompts via SAVPE (\cref{sec:savpe}), and prompt-free scenario with LRPC (\cref{sec:lrpc}).

\subsection{Model architecture}
\label{sec:architecture}
As shown in \cref{fig:pipeline}, YOLOE adopts the typical YOLOs' architecture~\cite{redmon2016you,bochkovskiy2020yolov4,jocherultralytics}, consisting of backbone, PAN, regression head, segmentation head, and object embedding head. The backbone and PAN extracts multi-scale features for the image. For each anchor point, the regression head predicts the bounding box for detection, and the segmentation head produces the prototype and mask coefficients for segmentation~\cite{bolya2019yolact}. The object embedding head follows the structure of classification head in YOLOs, except that the output channel number of last 1$\times$ convolution layer is changed from the class number in closed-set scenario to the embedding dimension. Meanwhile, given text and visual prompts, we employ RepRTA and SAVPE to encode them as normalized prompt embeddings $\mathcal{P}$, respectively. They serve as the classification weights and contrast with the anchor points' object embeddings $\mathcal{O}$ to obtain category labels. The process can be formalized as
\begin{equation}
    \small
    \setlength{\abovedisplayskip}{5pt}
    \setlength{\belowdisplayskip}{5pt}
    \text{Label} = \mathcal{O} \cdot \mathcal{P}^T: \mathbb{R}^{N\times D} \times \mathbb{R}^{D \times C} \rightarrow \mathbb{R}^{N \times C},
\end{equation} where $N$ denotes the number of anchor points, $C$ indicates the number of prompts, and $D$ means the feature dimension of embeddings, respectively. 

\subsection{Re-parameterizable region-text alignment}
\label{sec:reprta}
In open-set scenarios, the alignment between textual and object embeddings determines the accuracy of identified categories. Prior works usually introduce complex cross-modality fusion to improve the visual-textual representation for better alignment~\cite{cheng2024yoloworld,liu2024gdino}. However, these ways incur notable computational overhead, especially with large number of texts. Given this, we present Re-parameterizable Region-Text Alignment (RepRTA) strategy, which improves pretrained textual embeddings during training through the re-parameterizable lightweight auxiliary network. The alignment between textual and anchor points' object embeddings can be enhanced with zero inference and transferring cost.

\begin{figure}[t]
    \centering
    \includegraphics[width=1.0\columnwidth]{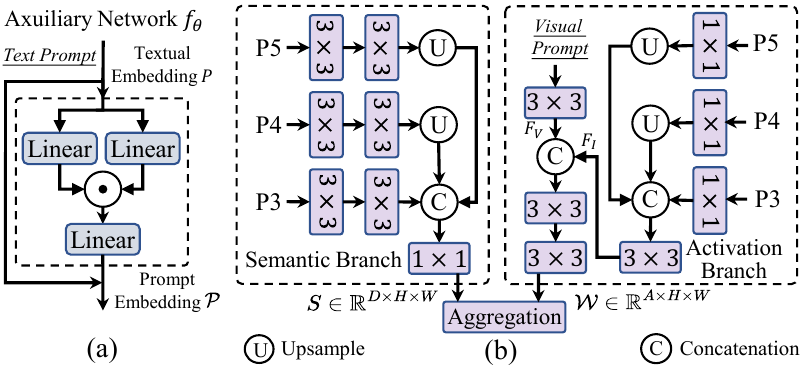}
    \caption{(a) The structure of lightweight auxiliary network in RepRTA, which consists of one SwiGLU FFN block~\cite{shazeer2020glu}. (b) The structure of SAVPE, which consists of semantic branch to generate prompt-agnostic semantic features and activation branch to provide grouped prompt-aware weights. Visual prompt embedding can thus be efficiently derived by their aggregation.}
    \label{fig:reprta}
    \vspace{-15pt}
\end{figure}

Specifically, with the text prompts of $T$ with length of $C$, we first employ the CLIP text encoder~\cite{radford2021clip,vasu2024mobileclip} to obtain pretrained textual embedding $P=\text{TextEncoder}(T)$. Before training, we cache all embeddings of texts in datasets in advance and the text encoder can be removed with no extra training cost. Meanwhile, as shown in \cref{fig:reprta}.(a), we introduce a lightweight auxiliary network $f_{\theta}$ with only one feed forward block~\cite{vaswani2017attention,shazeer2020glu}, where $\theta$ indicates the trainable parameters and introduces low overhead compared with closed-set training. It derives the enhanced textual embedding $\mathcal{P}=f_{\theta}(P)\in\mathbb{R}^{C \times D}$ for contrasting with the anchor points' object embedding during training, leading to improved visual-semantic alignment. Let $K \in \mathbb{R}^{D\times D' \times 1 \times 1}$ be the kernel parameters of last convolution layer with input features $I \in \mathbb{R}^{D' \times H \times W}$ in the object embedding head, $\circledast$ be the convolution operator, and $R$ be the reshape function, we have
\begin{equation}
    \small
    \setlength{\abovedisplayskip}{5pt}
    \setlength{\belowdisplayskip}{5pt}
    \text{Label} = R_{{D\times H \times W\rightarrow HW \times D} }(I \circledast K) \cdot (f_{\theta}(P))^T.
\end{equation} Moreover, after training, the auxiliary network can be re-parameterized with the object embedding head into the identical classification head of YOLOs. The new kernel parameters $K'\in \mathbb{R}^{C\times D' \times 1 \times 1}$ for last convolution layer after re-parameterization can be derived by
\begin{equation}
    \small
    \setlength{\abovedisplayskip}{5pt}
    \setlength{\belowdisplayskip}{5pt}
    K' = R_{C\times D \rightarrow C \times D \times 1 \times 1}  (f_{\theta}(P)) \circledast K^T.
\end{equation} The final predication can be obtained by $\text{Label}=I \circledast K'$, which is identical to the original YOLO architecture, leading to zero overhead for deployment and transferring to downstream closed-set tasks.

\subsection{Semantic-activated visual prompt encoder}
\label{sec:savpe}
Visual prompts are designed to indicate the object category of interest through visual cues, \eg, box and mask. To produce the visual prompt embedding, prior works often employ transformer-heavy design~\cite{jiang2024trex2,li2024dinov}, \eg, deformable attention~\cite{zhu2021deformabledetr}, or additional CLIP vision encoder~\cite{radford2021clip,zang2022ovdetr}. These ways, however, introduce challenges in deployment and efficiency due to complex operators or high computational demands. Considering this, we introduce Semantic-Activated Visual Prompt Encoder (SAVPE) for efficiently processing visual cues. It features two decoupled lightweight branches: (1) Semantic branch outputs prompt-agnostic semantic features in $D$ channels without overhead of fusing visual cues, and (2) Activation branch produces grouped prompt-aware weights by interacting visual cues with image features in much fewer channels under low costs. Their aggregation then leads to informative prompt embedding under minimal complexity.

As shown in \cref{fig:reprta}.(b), in the semantic branch, we adopt the similar structure as object embedding head. With multi-scale features $\{P_3, P_4, P_5\}$ from PAN, we employ two 3$\times$3 convs for each scale, respectively. After upsampling, features are concatenated and projected to derive semantic features $S \in \mathbb{R}^{D \times H \times W}$. In the activation branch, we formalize visual prompt as mask with 1 for indicated region and 0 for others. We downsample it and leverage 3$\times$3 conv to derive prompt feature $F_{V} \in \mathbb{R}^{A \times H \times W}$. Besides, we obtain image features $F_{I} \in \mathbb{R}^{A \times H \times W}$ for fusion with it from $\{P_3, P_4, P_5\}$ by convs. $F_{V}$ and $F_{I}$ are then concatenated and utilized to output prompt-aware weights $\mathcal{W} \in \mathbb{R}^{A \times H \times W}$, which is normalized using softmax within prompt-indicated region. Moreover, we divide the channels of $S$ into $A$ groups with $\frac{D}{A}$ channels in each. The channels in the $i$-th group share the weight $\mathcal{W}_{i:i+1}$ from the $i$-th channel of $\mathcal{W}$. With $A\ll D$, we can process visual cues with image features in low dimension, bringing minimal cost. Furthermore, prompt embedding can be derived with aggregation of two branches by
\begin{equation}
    \small
    \setlength{\abovedisplayskip}{5pt}
    \setlength{\belowdisplayskip}{5pt}
    \mathcal{P} = \text{Concat}(G_1, ..., G_A); G_i = \mathcal{W}_{i:i+1} \cdot S^T_{\frac{D}{A} * i:\frac{D}{A} * (i+1)}.
\end{equation} It can thus contrast with anchor points' object embeddings to identify objects with category of interest.

\subsection{Lazy region-prompt contrast}
\label{sec:lrpc}
In prompt-free scenario without explicit guidance, models are expected to identity all objects with names in the image. Prior works usually formulate such setting as a generative problem, where language model is employed to generate categories for dense found objects~\cite{lin2024generateu,ren2024dinox,wu2024grit}. However, this introduces notable overhead, where language models, \eg, FlanT5-base~\cite{chung2024scaling} with 250M parameters in GenerateU~\cite{lin2024generateu} and OPT-125M~\cite{zhang2022opt} in DINO-X~\cite{ren2024dinox}, are far from meeting high efficiency requirement. Given this, we reformulate such setting as a retrieval problem and present Lazy Region-Prompt Contrast (LRPC) strategy. It lazily retrieves category names from a built-in large vocabulary for anchor points with objects in the cost-effective way. Such paradigm enjoys zero dependency on language models, meanwhile with favorable efficiency and performance.

Specifically, with pretrained YOLOE, we introduce a specialized prompt embedding and train it exclusively to find all objects, where objects are treated as one category. Meanwhile, we follow~\cite{huang2023ram} to collect a large vocabulary which covers various categories and serve as the built-in data source for retrieval. One may directly leverage the large vocabulary as text prompts for YOLOE to identify all objects, which, however, incurs notable computational cost by contrasting abundant anchor points' object embeddings with numerous textual embeddings. Instead, we employ the specialized prompt embedding $\mathcal{P}_s$ to find the set $\mathcal{O}'$ of anchor points corresponding to objects by
\begin{equation}
    \small
    \setlength{\abovedisplayskip}{5pt}
    \setlength{\belowdisplayskip}{5pt}
    \mathcal{O}'=\{o \in \mathcal{O}\ |\ o \cdot \mathcal{P}_s^T > \delta\}, 
\end{equation} where $\mathcal{O}$ denotes all anchor points and $\delta$ is the threshold hyperparameter for filtering. Then, only anchor points in $\mathcal{O}'$ are lazily matched against the built-in vocabulary to retrieve category names, bypassing the cost for irrelevant anchor points. This further improves efficiency without performance drop, facilitating the real world application.

\subsection{Training objective}
During training, we follow~\cite{cheng2024yoloworld} to obtain an online vocabulary for each mosaic sample with the texts involved in the images as positive labels. Following~\cite{jocherultralytics}, we leverage task-aligned label assignment to match predictions with ground truths. The binary cross entropy loss is employed for classification, with IoU loss and distributed focal loss adopted for regression. For segmentation, we follow~\cite{bolya2019yolact} to utilize binary cross-entropy loss for optimizing masks.

\section{Experiments}
\subsection{Implementation details}
\textbf{Model.} For fair comparison with~\cite{cheng2024yoloworld}, we employ the same YOLOv8 architecture~\cite{jocherultralytics} for YOLOE. Besides, to verify its good generalizability on other YOLOs, we also experiment with YOLO11 architecture~\cite{jocherultralytics}. For both of them, we provide three model scales, \ie, small (S), medium (M), and large (L), to suit various application needs. Text prompts are encoded using the pretrained MobileCLIP-B(LT)~\cite{vasu2024mobileclip} text encoder. We empirically use $A=16$ in SAVPE, by default.

\textbf{Data.} We follow \cite{cheng2024yoloworld} to utilize detection and grounding datasets, including Objects365 (V1)~\cite{shao2019objects365}, GoldG~\cite{kamath2021mdetr} (includes GQA~\cite{hudson2019gqa} and Flickr30k~\cite{plummer2015flickr30k}), where images from COCO~\cite{lin2014mscoco} are excluded. Beside, we leverage advanced SAM-2.1 \cite{ravi2024sam2} model to generate pseudo instance masks using ground truth bounding boxes from the detection and grounding datasets for segmentation data. These masks undergo filtering and simplification to eliminate noise~\cite{douglas1973polydp}. For visual prompt data, we follow~\cite{jiang2024trex2} to leverage ground truth bounding boxes for visual cues. In prompt-free tasks, we reuse the same datasets, but annotate all objects as a single category to learn a specialized prompt embedding.

\textbf{Training.} Due to limited computational resource, unlike YOLO-World's training for 100 epochs, we first train YOLOE with text prompts for 30 epochs. Then, we only train the SAVPE for merely 2 epochs with visual prompts, which avoids additional significant training cost that comes with supporting visual prompts. At last, we train the specialized prompt embedding for only 1 epoch for prompt-free scenarios. During the text prompt training stage, we adopt the same settings as~\cite{cheng2024yoloworld}.
Notably, YOLOE-v8-S / M / L can be trained on 8 Nvidia RTX4090 GPUs in 12.0 / 17.0 / 22.5 hours, with 3$\times$ less cost compared with YOLO-World. For visual prompt training, we freeze all other parts and adopt the same setting as in text prompt training. To enable prompt-free capability, we leverage the same data to train a specialized embedding. We can see that YOLOE not only enjoys low training costs but also show exceptional zero-shot performance. Besides, to verify YOLOE's good transferability on downstream tasks, we fine-tune our YOLOE on COCO~\cite{lin2014mscoco} for closed-set detection and segmentation. We experiment with two distinct practical fine-tuning strategies: (1) \textit{Linear probing}: Only the classification head is learnable and (2) \textit{Full tuning}: All parameters are trainable. 
For \textit{Linear probing}, we train all models for only 10 epochs. For \textit{Full tuning}, we train small scale models including YOLOE-v8-S / 11-S for 160 epochs, and medium and large scale models including YOLOE-v8-M / L and YOLOE-11-M / L for 80 epochs, respectively.

\begin{table*}[t]
\centering
\caption{\textbf{Zero-shot detection evaluation on LVIS.} For fair comparisons, \textit{Fixed AP} is reported on LVIS \texttt{minival} set in a zero-shot manner. The training time is for text prompts, based on 8 Nvidia V100 GPUs for~\cite{li2022glip,yao2022detclip} and 8 RTX4090 GPUs for YOLO-World and YOLOE. The FPS is measured on Nvidia T4 GPU using TensorRT and on iPhone 12 using CoreML, respectively. Results are provided with text prompt (T) and visual prompt (V) type. For training data, OI, HT, and CH indicates OpenImages~\cite{krasin2017openimages}, HierText~\cite{long2022towards}, and CrowdHuman~\cite{shao2018crowdhuman}, respectively. OG indicates Objects365~\cite{shao2019objects365} and GoldG~\cite{kamath2021mdetr}, and G-20M represents Grounding-20M~\cite{ren2024groundingdino15advance}.}
\label{tab:lvis-det}
\small
\setlength{\tabcolsep}{5.5pt}
\begin{tabular}{l>{\hspace{-8pt}}c<{\hspace{-3pt}}c>{\hspace{-5pt}}c>{\hspace{-5pt}}ccccccc} 
\toprule
\multirow{2}{*}{Model} & \multirow{2}{*}{\makecell{Prompt\\Type}} & \multirow{2}{*}{Params} & \multirow{2}{*}{\makecell{Training\\Data}} & \multirow{2}{*}{\makecell{Training\\Time}} & \multirow{2}{*}{\makecell{FPS\\T4 / iPhone}} & \multirow{2}{*}{AP} & \multirow{2}{*}{AP$_r$} & \multirow{2}{*}{AP$_c$} & \multirow{2}{*}{AP$_f$}  \\
\\
\midrule 
GLIP-T~\cite{li2022glip} & T & 232M & OG,Cap4M & 1337.6h & - / - & 26.0 & 20.8 & 21.4 & 31.0 \\
GLIPv2-T~\cite{zhang2022glipv2} & T & 232M & OG,Cap4M & - & - / - & 29.0 & - & - & - \\
GDINO-T~\cite{liu2024gdino} & T & 172M & OG,Cap4M & - & - / - & 27.4 & 18.1 & 23.3 & 32.7 \\
DetCLIP-T~\cite{yao2022detclip} & T & 155M & OG & 250.0h & - / - & 34.4 & 26.9 & 33.9 & 36.3 \\
G-1.5 Edge~\cite{ren2024groundingdino15advance} & T & - & G-20M & - & - / - & 33.5 & 28.0 & 34.3 & 33.9 \\
T-Rex2~\cite{jiang2024trex2} & V & - & \makecell{O365,OI,HT\\CH,SA-1B} & - &  - / - & 37.4 & 29.9 & 33.9 & 41.8 \\ 
YWorldv2-S~\cite{cheng2024yoloworld} & T & 13M & OG & 41.7h & 216.4 / 48.9 & 24.4 & 17.1 & 22.5 & 27.3 \\
YWorldv2-M~\cite{cheng2024yoloworld} & T & 29M & OG & 60.0h & 117.9 / 34.2 & 32.4 & 28.4 & 29.6 & 35.5 \\
YWorldv2-L~\cite{cheng2024yoloworld} & T & 48M & OG & 80.0h & ~~80.0 / 22.1 & 35.5 & 25.6 & 34.6 & 38.1 \\
\midrule
\textbf{YOLOE-v8-S} & T / V & 12M / 13M & OG & \textbf{12.0h} & \textbf{305.8} / \textbf{64.3} & \textbf{27.9} / 26.2 & \textbf{22.3} / 21.3 & \textbf{27.8} / 27.7 & \textbf{29.0} / 25.7 \\
\textbf{YOLOE-v8-M} & T / V & 27M / 30M & OG & \textbf{17.0h} & \textbf{156.7} / \textbf{41.7} & \textbf{32.6} / 31.0 & 26.9 / \textbf{27.0} & \textbf{31.9} / 31.7 & \textbf{34.4} / 31.1 \\
\textbf{YOLOE-v8-L} & T / V & 45M / 50M & OG & \textbf{22.5h} & \textbf{102.5} / \textbf{27.2} & \textbf{35.9} / 34.2 & \textbf{33.2} / \textbf{33.2} & \textbf{34.8} / 34.6 & \textbf{37.3} / 34.1 \\
\textbf{YOLOE-11-S} & T / V & 10M / 12M & OG & \textbf{13.0h} & \textbf{301.2} / \textbf{73.3} & \textbf{27.5} / 26.3 & 21.4 / \textbf{22.5} & 26.8 / \textbf{27.1} & \textbf{29.3} / 26.4 \\
\textbf{YOLOE-11-M} & T / V & 21M / 27M & OG & \textbf{18.5h} & \textbf{168.3} / \textbf{39.2} & \textbf{33.0} / 31.4 & 26.9 / \textbf{27.1} & \textbf{32.5} / 31.9 & \textbf{34.5} / 31.7 \\
\textbf{YOLOE-11-L} & T / V & 26M / 32M & OG & \textbf{23.5h} & \textbf{130.5} / \textbf{35.1} & \textbf{35.2} / 33.7 & \textbf{29.1} / 28.1 & \textbf{35.0} / 34.6 & \textbf{36.5} / 33.8 \\ 
\bottomrule
\end{tabular}
\vspace{-10pt}
\end{table*}

\textbf{Metric.} For text prompt evaluation, we utilize all category names from the benchmark as inputs, adhering to the standard protocol for open-vocabulary object detection tasks. For visual prompt evaluation, following~\cite{jiang2024trex2}, for each category, we randomly sample $N$ training images ($N$=16 by default), extract visual embeddings using their ground truth bounding boxes, and compute the average prompt embedding. For prompt-free evaluation, we employ the same protocol as~\cite{lin2024generateu}. A pretrained text encoder~\cite{vasu2024mobileclip} is employed to map open-ended predictions to semantically similar category names within the benchmark. In contrast to~\cite{lin2024generateu}, we streamline the mapping process by selecting the most confident prediction, and eliminating the need for top-k selection and beam search. We use the tag list from~\cite{huang2023ram} as the built-in large vocabulary with total 4585 category names, and empirically use $\delta=0.001$ for LRPC, by default. For all three prompt types, following~\cite{cheng2024yoloworld,jiang2024trex2,lin2024generateu}, evaluations are conducted on LVIS~\cite{gupta2019lvis} in a zero-shot manner, which contains 1,203 categories. By default, \textit{Fixed AP}~\cite{dave2021evaluating} on LVIS \texttt{minival} subset is reported. For transferring to COCO, standard AP is evaluated, following~\cite{bochkovskiy2020yolov4,jocherultralytics}. Besides, we measure the FPS for all models on Nvidia T4 GPU with TensorRT and mobile device iPhone 12 with CoreML.

\subsection{Text and visual prompt evaluation}
As shown in \cref{tab:lvis-det}, for detection on LVIS, YOLOE exhibits favorable trade-offs between efficiency and zero-shot performance across different model scales. We also note that such results are achieved under much less training time, \eg, 3$\times$ faster than YOLO-Worldv2. Specifically, YOLOE-v8-S / M / L outperforms YOLOv8-Worldv2-S / M / L by 3.5 / 0.2 / 0.4 AP, along with 1.4$\times$ / 1.3$\times$ / 1.3$\times$ and 1.3$\times$ / 1.2$\times$ / 1.2$\times$ inference speedups on T4 and iPhone 12, respectively. Besides, for rare category which is challenging, our YOLOE-v8-S and YOLOE-v8-L obtains significant improvements of 5.2\% and 7.6\% AP$_r$. Besides, compared with YOLO-Worldv2, while YOLOE-v8-M / L achieves lower AP$_f$, this performance gap primarily stems from YOLOE's integration of both detection and segmentation in one model. Such multi-task learning introduces a trade-off that adversely impact detection performance on frequent categories, as shown in \cref{tab:roadmap}. Besides, YOLOE with YOLO11 architecture also exhibits favorable performance and efficiency. For example, YOLOE-11-L achieves comparable AP with YOLO-Worldv2-L, but with notably 1.6$\times$ inference speedups on T4 and iPhone 12, highlighting the strong generalizability of our YOLOE. 

Moreover, the inclusion of visual prompts further amplifies YOLOE's versatility. Compared with T-Rex2, YOLOE-v8-L yield the improvements of 3.3 AP$_r$ and 0.9 AP$_c$, with 2$\times$ less training data (3.1 M vs. Our: 1.4 M) and much lower training resource (16 Nvidia A100 GPUs vs. Our: 8 Nvidia RTX4090 GPUs). Besides, for visual prompts, while we only train SAVPE with other parts frozen for 2 epochs, we note that it can achieve comparable AP$_r$ and AP$_c$ with the text prompts for various model scales. This shows the efficacy of visual prompts in less frequent objects that text prompts often struggle to accurately describe, which is similar to the observation in \cite{jiang2024trex2}.

\begin{table}[t]
\centering
\caption{\textbf{Segmentation evaluation on LVIS.} We evaluate all models on LVIS \texttt{val} set with the standard AP$^m$ reported. YOLOE supports both text (T) and visual cues (V) as inputs. $\dagger$ indicates that the pretrained models are fine-tuned on \texttt{LVIS-Base} data for segmentation head. In contrast, we evaluate YOLOE in a zero-shot manner without utilizing any images from LVIS during training.}
\label{tab:lvis-segm}
\small
\setlength{\tabcolsep}{2.5pt}
\resizebox{\linewidth}{!}{%
\begin{tabular}{lccccccc} 
\toprule
Model & Prompt & AP$^m$ & AP$_r^m$ & AP$_c^m$ & AP$_f^m$ \\ 
\midrule
YWorld-M$^\dagger$ & T & 16.7 & 12.6 & 14.6 & 20.8  \\
YWorld-L$^\dagger$ & T & 19.1 & 14.2 & 17.2 & 23.5 \\
YWorldv2-M$^\dagger$ & T & 17.8 & 13.9 & 15.5 & 22.0 \\
YWorldv2-L$^\dagger$ & T & 19.8 & 17.2 & 17.5 & 23.6 \\
\midrule
\textbf{YOLOE-v8-S} & T / V & \textbf{17.7} / 16.8 & \textbf{15.5} / 13.5 & 16.3 / \textbf{16.7} & \textbf{20.3} / 18.2  \\
\textbf{YOLOE-v8-M} & T / V & \textbf{20.8} / 20.3 & \textbf{17.2} / 17.0 & 19.2 / \textbf{20.1} & \textbf{24.2} / 22.0 \\
\textbf{YOLOE-v8-L} & T / V & \textbf{23.5} / 22.0 & \textbf{21.9} / 16.5 & 21.6 / \textbf{22.1} & \textbf{26.4} / 24.3 \\
\textbf{YOLOE-11-S} & T / V & \textbf{17.6} / 17.1 & \textbf{16.1} / 14.4 & 15.6 / \textbf{16.8} & \textbf{20.5} / 18.6 \\
\textbf{YOLOE-11-M} & T / V & \textbf{21.1} / 21.0 & 17.2 / \textbf{18.3} & 19.6 / \textbf{20.6} & \textbf{24.4} / 22.6 \\
\textbf{YOLOE-11-L} & T / V & \textbf{22.6} / 22.5 & 19.3 / \textbf{20.5} & 20.9 / \textbf{21.7} & \textbf{26.0} / 24.1 \\
\bottomrule
\end{tabular}
}
\vspace{-5pt}
\end{table}

Furthermore, for segmentation, we present the evaluation results on the LVIS \texttt{val} set with the standard AP$^m$ reported in \cref{tab:lvis-segm}. It shows that YOLOE exhibits strong performance by leveraging both text prompts and visual prompts. Specifically, YOLOE-v8-M / L achieves 20.8 and 23.5 AP$^m$ in the zero-shot manner, significantly outperforming YOLO-Worldv2-M / L that is fine-tuned on \texttt{LVIS-Base} dataset, by 3.0 and 3.7 AP$^m$, respectively. These results well show the superiority of YOLOE.

\subsection{Prompt-free evaluation}
As shown in \cref{tab:lvis-pf}, for prompt-free scenario, YOLOE also exhibits superior performance and efficiency. Specifically, YOLO-v8-L achieves 27.2 AP and 23.5 AP$_r$, outperforming GenerateU with Swin-T backbone by 0.4 AP and 3.5 AP$_r$, along with 6.3$\times$ fewer parameters and 53$\times$ inference speedups. It shows the effectiveness of YOLOE by reformulating the open-ended problem as the retrieval task for a built-in large vocabulary and underscores its potential in generalizing across a wide range of categories without replying on explicit prompts. Such functionality also enhances YOLOE's practicality, enabling its application in a broader range of real-world scenarios.

\begin{table}[t]
\centering
\caption{\textbf{Prompt-free evaluation on LVIS.} \textit{Fixed AP} is reported on the LVIS \texttt{minival} set, following the protocol in \cite{lin2024generateu}. The FPS is measured on Nvidia T4 GPU with Pytorch~\cite{paszke2019pytorch}.}
\label{tab:lvis-pf}
\small
\setlength{\tabcolsep}{2pt}
\begin{tabular}{lccccccc} 
\toprule
Model & Backbone & Params & AP & AP$_r$ & AP$_c$ & AP$_f$ & FPS  \\ 
\midrule
GenerateU~\cite{lin2024generateu} & Swin-T & 297M & 26.8 & 20.0 & 24.9 & 29.8 & 0.48 \\ 
GenerateU~\cite{lin2024generateu} & Swin-L & 467M & 27.9 & 22.3 & 25.2 & 31.4 & 0.40 \\
\midrule
\textbf{YOLOE-v8-S} & YOLOv8-S & 13M & \textbf{21.0} & 19.1 & 21.3 & 21.0 & \textbf{95.8} \\
\textbf{YOLOE-v8-M} & YOLOv8-M & 29M & \textbf{24.7} & 22.2 & 24.5 & 25.3 & \textbf{45.9} \\
\textbf{YOLOE-v8-L} & YOLOv8-L & 47M & \textbf{27.2} & 23.5 & 27.0 & 28.0 & \textbf{25.3}  \\
\textbf{YOLOE-11-S} & YOLO11-S & 11M & \textbf{20.6} & 18.4 & 20.2 & 21.3 & \textbf{93.0} \\
\textbf{YOLOE-11-M} & YOLO11-M & 24M & \textbf{25.5} & 21.6 & 25.5 & 26.1 & \textbf{42.5} \\
\textbf{YOLOE-11-L} & YOLO11-L & 29M & \textbf{26.3} & 22.7 & 25.8 & 27.5 & \textbf{34.9}  \\
\bottomrule
\end{tabular}
\vspace{-10pt}
\end{table}

\subsection{Downstream transferring}
As shown in \cref{tab:coco-tf}, when transferring to COCO for downstream closed-set detection and segmentation, YOLOE exhibits favorable performance under limited training epochs in both two fine-tuning strategies. Specifically, for \textit{Linear probing}, with less than 2\% of the training time, YOLOE-11-M / L can achieve over 80\% of the performance of YOLO11-M / L, respectively. This highlights the strong transferability of YOLOE. For \textit{Full tuning}, YOLOE can further enhance the performance under limited training cost. For example, with nearly 4$\times$ less training epochs, YOLOE-v8-M / L outperforms YOLOv8-M / L by 0.4 AP$^m$ and 0.6 AP$^b$, respectively. Under 3$\times$ less training time, YOLO-v8-S also obtains better performance compared with YOLOv8-S for both detection and segmentation. These results well demonstrate that YOLOE can serve as a strong starting point for transferring to downstream task.

\begin{table}[t]
\centering
\caption{\textbf{Downstream transfer on COCO.} We fine-tune YOLOE on COCO and report the standard AP for both detection and segmentation. We experiment with two practical fine-tuning strategies, \ie, \textit{Linear probing} and \textit{Full tuning}.}
\label{tab:coco-tf}
\small
\setlength{\tabcolsep}{3pt}
\begin{tabular}{lccccccc} 
\toprule
Model & Epochs & AP$^b$ & AP$^b_{50}$ & AP$^b_{75}$ & AP$^m$ & AP$^m_{50}$ & AP$^m_{75}$ \\ 
\hline
\multicolumn{7}{c}{\textit{Training from scratch}} \\
YOLOv8-S & 500  & 44.7 & 61.4 & 48.7 & 36.6 & 58.0 & 38.6  \\
YOLOv8-M & 300 & 50.0 & 66.8 & 54.8 & 40.5 & 63.4 & 43.3  \\
YOLOv8-L & 300 & 52.4 & 69.3 & 57.2 & 42.3 & 66.0 & 44.9  \\
YOLO11-S & 500 & 46.6 & 63.3 & 50.6 & 37.8 & 59.7 & 40.0  \\
YOLO11-M & 600 & 51.5 & 68.5 & 55.7 & 41.5 & 65.0 & 43.9  \\
YOLO11-L & 600 & 53.3 & 70.1 & 58.2 & 42.8 & 66.8 & 45.5  \\ 
\hline
\multicolumn{7}{c}{\textit{Linear probing}} \\
\textbf{YOLOE-v8-S} & \textbf{10} & \textbf{35.6} & 51.5 & 38.9 & \textbf{30.3} & 48.2 & 32.0 \\
\textbf{YOLOE-v8-M} & \textbf{10} & \textbf{42.2} & 59.2 & 46.3 & \textbf{35.5} & 55.6 & 37.7 \\ 
\textbf{YOLOE-v8-L} & \textbf{10} & \textbf{45.4} & 63.3 & 50.0 & \textbf{38.3} & 59.6 & 40.8 \\
\textbf{YOLOE-11-S} & \textbf{10} & \textbf{37.0} & 52.9 & 40.4 & \textbf{31.5} & 49.7 & 33.5 \\
\textbf{YOLOE-11-M} & \textbf{10} & \textbf{43.1} & 60.6 & 47.4 & \textbf{36.5} & 56.9 & 39.0 \\
\textbf{YOLOE-11-L} & \textbf{10} & \textbf{45.1} & 62.8 & 49.5 & \textbf{38.0} & 59.2 & 40.6 \\ 
\hline
\multicolumn{7}{c}{\textit{Full tuning}} \\
\textbf{YOLOE-v8-S} & \textbf{160} & \textbf{45.0} & 61.6 & 49.1 & \textbf{36.7} & 58.3 & 39.1 \\
\textbf{YOLOE-v8-M} & \textbf{80} & \textbf{50.4} & 67.0 & 55.2 & \textbf{40.9} & 63.7 & 43.5 \\
\textbf{YOLOE-v8-L} & \textbf{80} & \textbf{53.0} & 69.8 & 57.9 & \textbf{42.7} & 66.5 & 45.6 \\
\textbf{YOLOE-11-S} & \textbf{160} & \textbf{46.2} & 62.9 & 50.0 & \textbf{37.6} & 59.3 & 40.1 \\
\textbf{YOLOE-11-M} & \textbf{80} & \textbf{51.3} & 68.3 & 56.0 & \textbf{41.5} & 64.8 & 44.3 \\
\textbf{YOLOE-11-L} & \textbf{80} & \textbf{52.6} & 69.7 & 57.5 & \textbf{42.4} & 66.2 & 45.2 \\
\bottomrule
\end{tabular}
\vspace{-5pt}
\end{table}

\subsection{Ablation study}
We further provide extensive analyses for the effectiveness of designs in our YOLOE. Experiments are conducted on YOLOE-v8-L and standard AP is reported on LVIS \texttt{minival} set for zero-shot evaluation, by default. 

\begin{table}[t]
\caption{Roadmap to YOLOE in terms of text prompts. The standard AP is reported on LVIS \texttt{minival} set in the zero-shot manner. The FPS is is measured on Nvidia T4 GPU and iPhone 12 with TensorRT (T) and CoreML (C), respectively.}
\label{tab:roadmap}
\centering
\small
\setlength{\tabcolsep}{2pt}
\begin{tabular}{l@{\hspace{-2pt}}ccccccccc}
\toprule
Model & Epochs & AP & AP$_r$ & AP$_c$ & AP$_f$ & FPS (T / C) \\
\toprule
YOLO-Worldv2-L & 100 & 33.0 & 22.6 & 32.0 &	35.8 & ~~80.0 / 22.1 \\
+ Fewer train. epochs & 30 & 31.0 & 22.6 & 28.8 & 34.2 & ~~80.0 / 22.1 \\
+ Global negative dict. & 30 & 31.9 & 22.8 & 31.0 & 34.4 & ~~80.0 / 22.1 \\
- Cross-modal. fusion & 30 & 30.0 & 19.1 & 28.0 & 33.9 & 102.5 / 27.2 \\
+ MobileCLIP encoder & 30 & 31.5 & 20.2 & 30.5 & 34.4 & 102.5 / 27.2 \\
+ RepRTA & 30 & 33.5 & 29.5 & 32.0 & 35.5 & 102.5 / 27.2 \\
+ Segment. (YOLOE) & 30 & 33.3 & 30.8 & 32.2 & 34.6 & 102.5 / 27.2 \\
\bottomrule
\end{tabular}
\vspace{-10pt}
\end{table}

\begin{figure*}[t]
    \centering
    \includegraphics[width=0.82\textwidth]{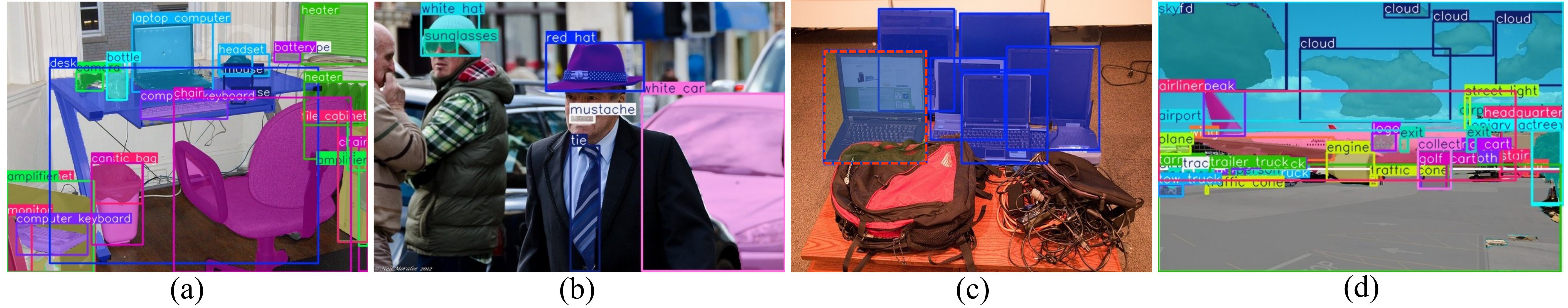}
    \caption{(a) Zero-shot inference on LVIS. (b) Results with customized text prompt, where ``white hat, red hat, white car, sunglasses, mustache, tie'' are provided as text prompts. (c) Results with visual prompt, where the red dashed bounding box serves as the visual cues. (d) Results in prompt-free scenario, where no explicit prompt is provided. Please refer to the supplementary for more examples.}
    \label{fig:visualization}
    \vspace{-13pt}
\end{figure*}

\textbf{Roadmap to YOLOE.} We outline the stepwise progression from the baseline model YOLOv8-Worldv2-L to our YOLOE-v8-L in terms of text prompts in \cref{tab:roadmap}. With the initial baseline metric of 33.0\% AP, due to limited computational resource, we first reduce the training epochs to 30, leading to 31.0\% AP. Besides, instead of using empty string as negative texts for grounding data, we follow~\cite{yao2022detclip} by maintaining a global dictionary to sample more diverse negative prompts. The global dictionary is constructed by selecting category names that appear more than 100 times in the training data. This leads to 0.9\% AP improvement. Next, we remove the cross-modality fusion to avoid costly visual-textual feature interaction, which results in 1.9\% AP degradation but with 1.28$\times$ and 1.23$\times$ inference speedups on T4 and iPhone 12, respectively. To address this drop, we utilize stronger MobileCLIP-B(LT) text encoder~\cite{vasu2024mobileclip} to obtain better pretrained textual embeddings, which recovers AP to 31.5\%. Furthermore, we employ RepRTA to enhance the alignment between anchor points' object and textual embeddings, which leads to notable 2.3\% AP enhancement with zero inference overhead, showing its effectiveness. At last, we introduce the segmentation head and train YOLOE for detection and segmentation simultaneously. Although this leads to 0.2\% AP and 0.9 AP$_f$ drop due to multi-task learning, YOLOE gains ability to segment arbitrary objects.

\begin{figure}[t]
\begin{minipage}{.216\textwidth}
    \captionof{table}{Effective. of SAVPE.}
    \label{tab:savpe}
    \centering
    \small
    \setlength{\tabcolsep}{2.5pt}
    \resizebox{1\columnwidth}{!}{
    \begin{tabular}{lccccc}
    \toprule
    Model & AP & AP$_r$ & AP$_c$ & AP$_f$ \\
    \toprule 
    Mask pool & 30.4 & 27.6 & 31.3 & 30.2 \\
    SAVPE & 31.9 & 29.4 & 32.5 & 31.7 \\
    \midrule
    $A=1$ & 30.9  & 28.2 & 31.9 & 30.4 \\
    $A=16$ & 31.9 & 29.4 & 32.5 & 31.7 \\
    $A=32$ & 31.9 & 28.2 & 33.0 & 31.7 \\
    \bottomrule
    \end{tabular}
    }
\end{minipage}
\hfill
\begin{minipage}{.257\textwidth}
    \captionof{table}{Effective. of LRPC.}
    \label{tab:lrpc}
    \centering
    \small
    \setlength{\tabcolsep}{2.45pt}
    \resizebox{1\columnwidth}{!}{
    \begin{tabular}{lcccccc}
    \toprule
    Model & LRPC & AP & AP$_r$ & AP$_c$ & AP$_f$ & FPS \\
    \toprule 
    \multirow{4}{*}{v8-S} & \ding{55} & 21.0 & 19.1 & 21.4 & 21.0 & 56.5 \\
     & $\delta=1e^{-3}$ & 21.0 & 19.1 & 21.3 & 21.0 & 95.8 \\
     & $\delta=1e^{-4}$ & 21.0 & 19.1 & 21.3 & 21.0 & 66.1 \\
     & $\delta=1e^{-2}$ & 20.8 & 19.1 & 21.2 & 20.8 & 106 \\
     \midrule
    \multirow{2}{*}{v8-L} & \ding{55} & 27.2 & 23.5 & 27.0 & 28.0 & 19.9 \\
     & $\delta=1e^{-3}$ & 27.2 & 23.5 & 27.0 & 28.0 & 25.3 \\
    \bottomrule
    \end{tabular}
    }
\end{minipage}
\vspace{-15pt}
\end{figure}

\textbf{Effectiveness of SAVPE.} To verify the effectiveness of SAVPE for visual inputs, we remove the activation branch and simply leverage mask pooling to aggregate semantic features with the formulated visual prompt mask. As shown in \cref{tab:savpe}, SAVPE significantly outperforms ``Mask pool'' by 1.5 AP. This is because ``Mask pool'' neglects the varying semantic importance at different positions within prompt-indicated region, while our activation branch effectively models such difference, leading to improved aggregation of semantic features and better prompt embedding for contrast. We also examine the impact of different group numbers, \ie, $A$, in the activation branch. As shown in \cref{tab:savpe}, performance can also be enhanced with only a group, \ie, $A=1$. Besides, we can achieve the strong performance of 31.9 AP under $A=16$, obtaining the favorable balance, where more groups lead to marginal performance difference.

\textbf{Effectiveness of LRPC.} To verify the effectiveness of LRPC for prompt-free setting, we introduce the baseline that directly leverage the built-in large vocabulary as text prompts for YOLOE to identify all objects. \cref{tab:lrpc} presents the comparison results. We observe that with the same performance, our LRPC obtains notably 1.7$\times$ / 1.3$\times$ inference speedups for YOLOE-v8-S / L, respectively, by lazily retrieving the categories for anchor points with found objects and skipping the numerous irrelevant ones. These results well highlight its efficacy and practicality. Besides, with different threshold $\delta$ for filtering, LRPC can achieve different performance and efficiency trade-offs, \eg, enabling 1.9$\times$ speedup for YOLOE-v8-S with only 0.2 AP drop.

\subsection{Visualization analyses}
We conduct visualization analyses for YOLOE in four scenarios: (1) Zero-shot inference on LVIS in \cref{fig:visualization}.(a), where its category names are text prompts, (2) Text prompts in \cref{fig:visualization}.(b), where arbitrary texts can be input as prompts, (3) Visual prompts in \cref{fig:visualization}.(c), where visual cues can be drawn as prompts, and (4) No explicit prompt in \cref{fig:visualization}.(d), where model identifies all objects. We can see that YOLOE performs well and can accurately detect and segment various objects in these diverse scenarios, further showing its efficacy and practicality in various applications.

\section{Conclusion}
In this paper, we present YOLOE, a single highly efficient model that seamlessly integrates object detection and segmentation across diverse open prompt mechanisms. Specifically, we introduce RepRTA, SAVPE, and LRPC to enable YOLOs to process textual prompt, visual cues, and prompt-free paradigm with favorable performance and low cost. Thanks to them, YOLOE enjoys strong capabilities and high efficiency for various prompt ways, enabling real-time seeing anything. We hope that it can serve as a strong baseline to inspire further advancements.

\section{Acknowledgments}
This work was supported by National Natural Science Foundation of China (Nos. 62525103, 624B2082, 62271281, 62441235).

{
    \small
    \bibliographystyle{ieeenat_fullname}
    \bibliography{main}
}

\appendix

\section{More Implementation Details}
\textbf{Data.} We employ Objects365\cite{shao2019objects365}, GoldG \cite{kamath2021mdetr} (including GQA\cite{hudson2019gqa} and Flickr30k \cite{plummer2015flickr30k}) for training YOLOE. \cref{tab:dataset} present their details. We utilize SAM-2.1-Hiera-Large~\cite{ravi2024sam2} to generate high-quality pseudo labeling of segmentation masks with ground truth bounding boxes as prompts. We filter out ones with too few areas. To enhance the smoothness of mask edges, we apply Gaussian kernel to masks, using 3$\times$3 and 7$\times$7 kernels for small and large ones, respectively. Besides, we refine the masks following~\cite{douglas1973polydp}, which iteratively simplifies the mask contours. This reduces noise pixels while preserving overall structure.

\begin{table}[h]
\centering
\caption{Data details for YOLOE training.}
\label{tab:dataset}
\small
\setlength{\tabcolsep}{3pt}
\begin{tabular}{llcccc} 
\toprule
Dataset & Type & Box & Mask & Images & Anno. \\ 
\midrule
Objects365~\cite{shao2019objects365} & Detection & $\checkmark$ & $\checkmark$ & 609k & 8,530k \\
GQA~\cite{hudson2019gqa} & Grounding & $\checkmark$ & $\checkmark$ & 621k & 3,662k \\
Flickr~\cite{plummer2015flickr30k} & Grounding & $\checkmark$ & $\checkmark$ & 149k & 638k \\
\bottomrule
\end{tabular}
\end{table}

\textbf{Training.} For all models, we adopt AdamW optimizer with an initial learning rate of 0.002. The batch size and weight decay are set to 128 and 0.025, respectively. The data augmentation includes color jittering, random affine transformations, random horizontal flipping, and mosaic augmentation. During transferring to COCO, for both \textit{Linear probing} and \textit{Full tuning}, we utilize the AdamW optimizer with an initial learning rate of 0.001, setting the batch size and weight decay to 128 and 0.025, respectively.

\section{More Analyses for LRPC}
\begin{figure}[t]
    \centering
    \includegraphics[width=0.45\textwidth]{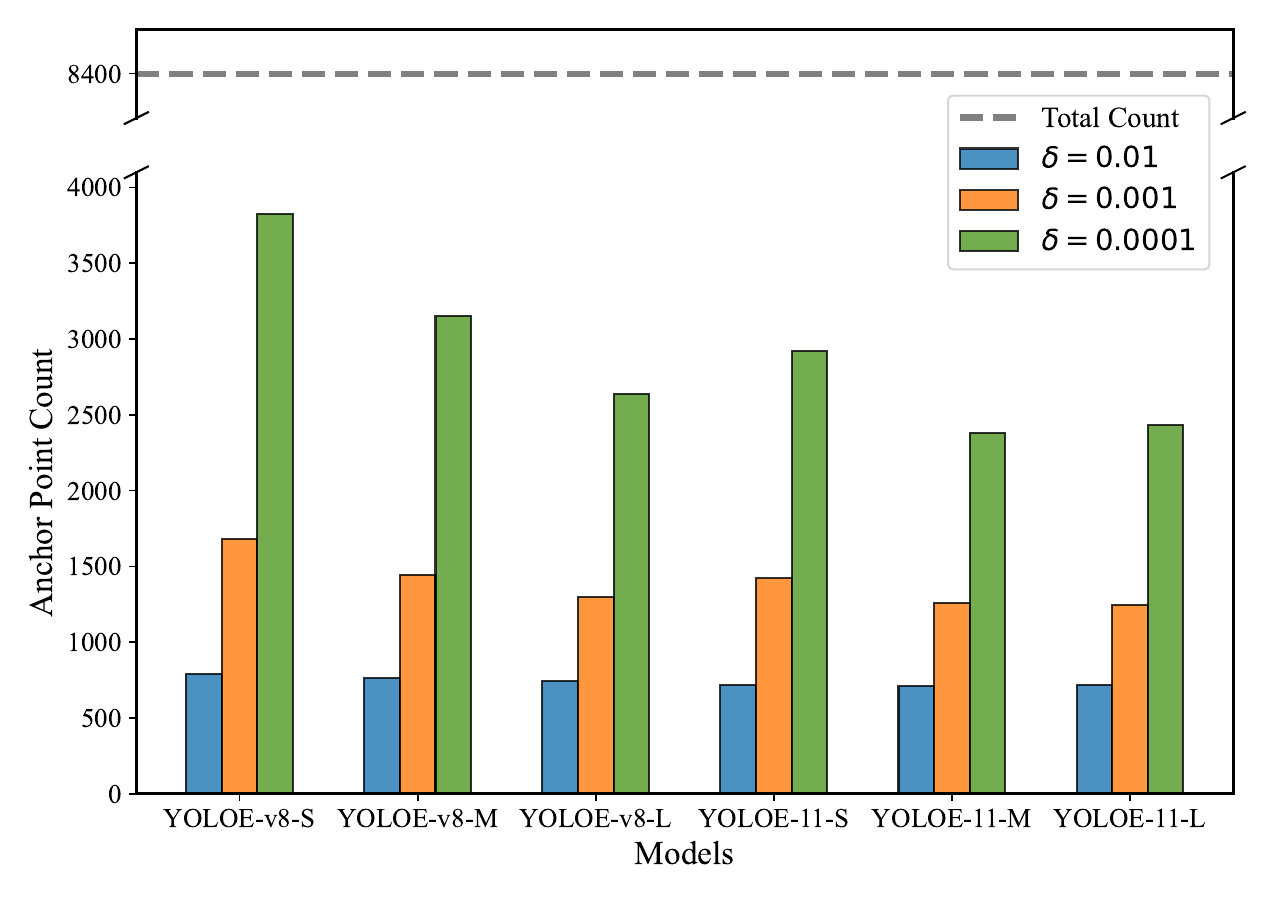}
    \caption{The count of retained anchor points under different filtering thresholds in LRPC. The dashed line means the total number.}
    \label{fig:lrpc-anchor}
    \vspace{-10pt}
\end{figure}

To qualitatively show the efficacy of LRPC strategy, we visualize the number of anchor points retained for category retrieval after filtering. We present their average count under varying filtering threshold $\delta$ on the LVIS \texttt{minival} set in \cref{fig:lrpc-anchor}. It reveals that as $\delta$ increases, the number of retained anchor points decreases substantially across different models. This reduction significantly lowers computational overhead compared with the baseline scenario, which employs a total of 8400 anchor points. For example, for YOLOE-v8-S, with $\delta=0.001$, the number of valid anchor points is reduced by 80\%, enjoying 1.7$\times$ inference speedup with the same performance (see Tab. 7 in the paper). The results further confirm the notably redundancy of anchor points for category retrieval and verify the efficacy of LRPC.

\section{More Visualization Results}
To qualitatively show the efficacy of YOLOE, we present more visualization results for it in various scenarios.

\begin{figure*}[t]
    \centering
    \includegraphics[width=0.85\textwidth]{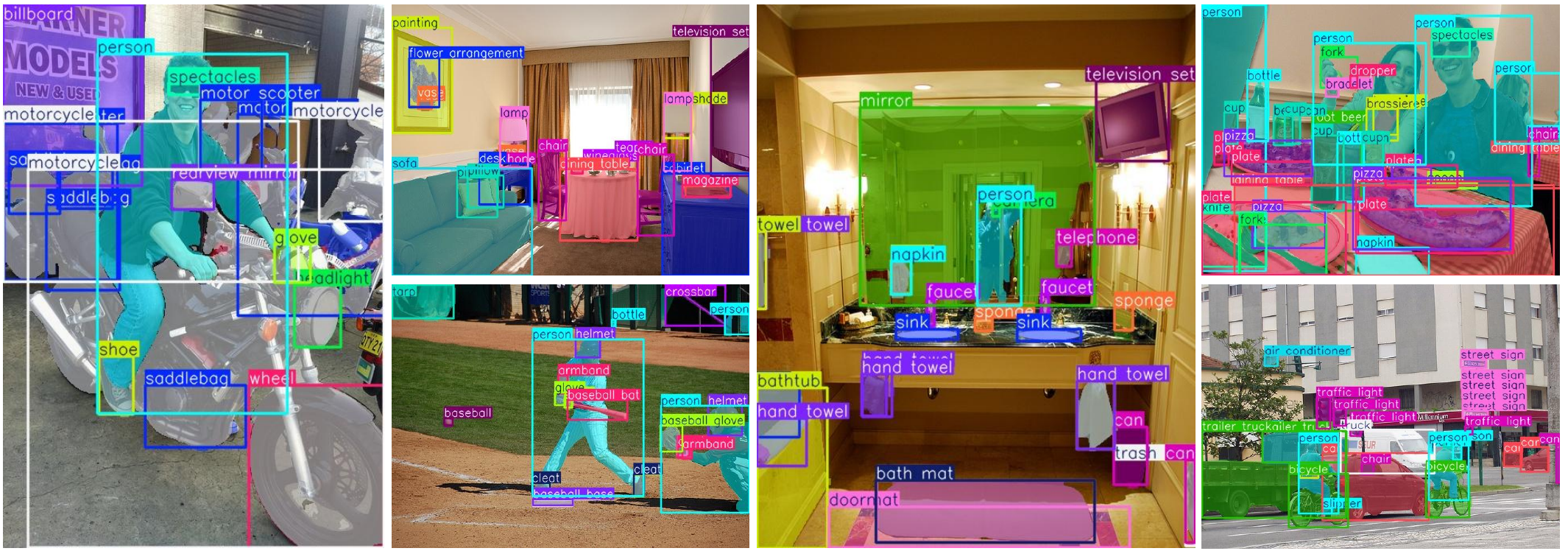}
    \caption{Zero-Shot inference on LVIS. The categories of LVIS are provided as text prompts.}
    \label{fig:tp-visual}
    \vspace{-10pt}
\end{figure*}

\textbf{Zero-shot inference on LVIS.} In \cref{fig:tp-visual}, we present the zero-shot inference capabilities of YOLOE on the LVIS. By leveraging the 1203 category names as text prompts, the model demonstrates its ability to detect and segment diverse objects across various images.

\begin{figure*}[t]
    \centering
    \includegraphics[width=0.85\textwidth]{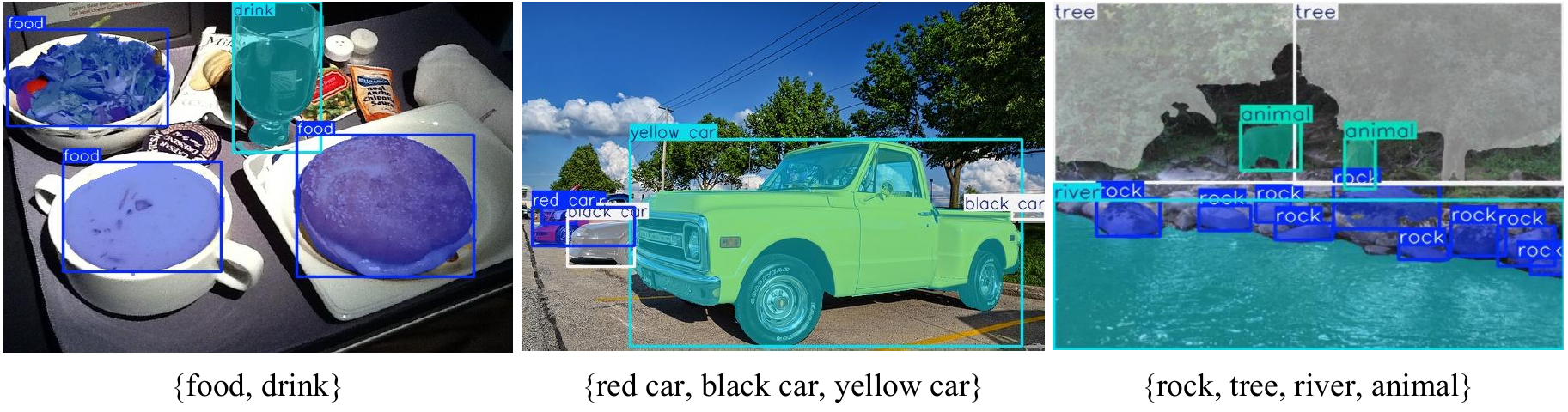}
    \caption{Prompt with customized texts. YOLOE adapts to both generic and specific text prompts for flexible usage.}
    \label{fig:cp-visual}
    \vspace{-10pt}
\end{figure*}

\textbf{Prompt with customized texts.} \cref{fig:cp-visual} presents the results with customized text prompts. We can see that YOLOE can interpret both generic and specific textual inputs, enabling precise object detection and fine-grained segmentation. Such capability allows users to tailor the model's behavior to meet specific requirements by defining input prompts at varying levels of granularity.

\begin{figure*}[!h]
    \centering
    \includegraphics[width=0.85\textwidth]{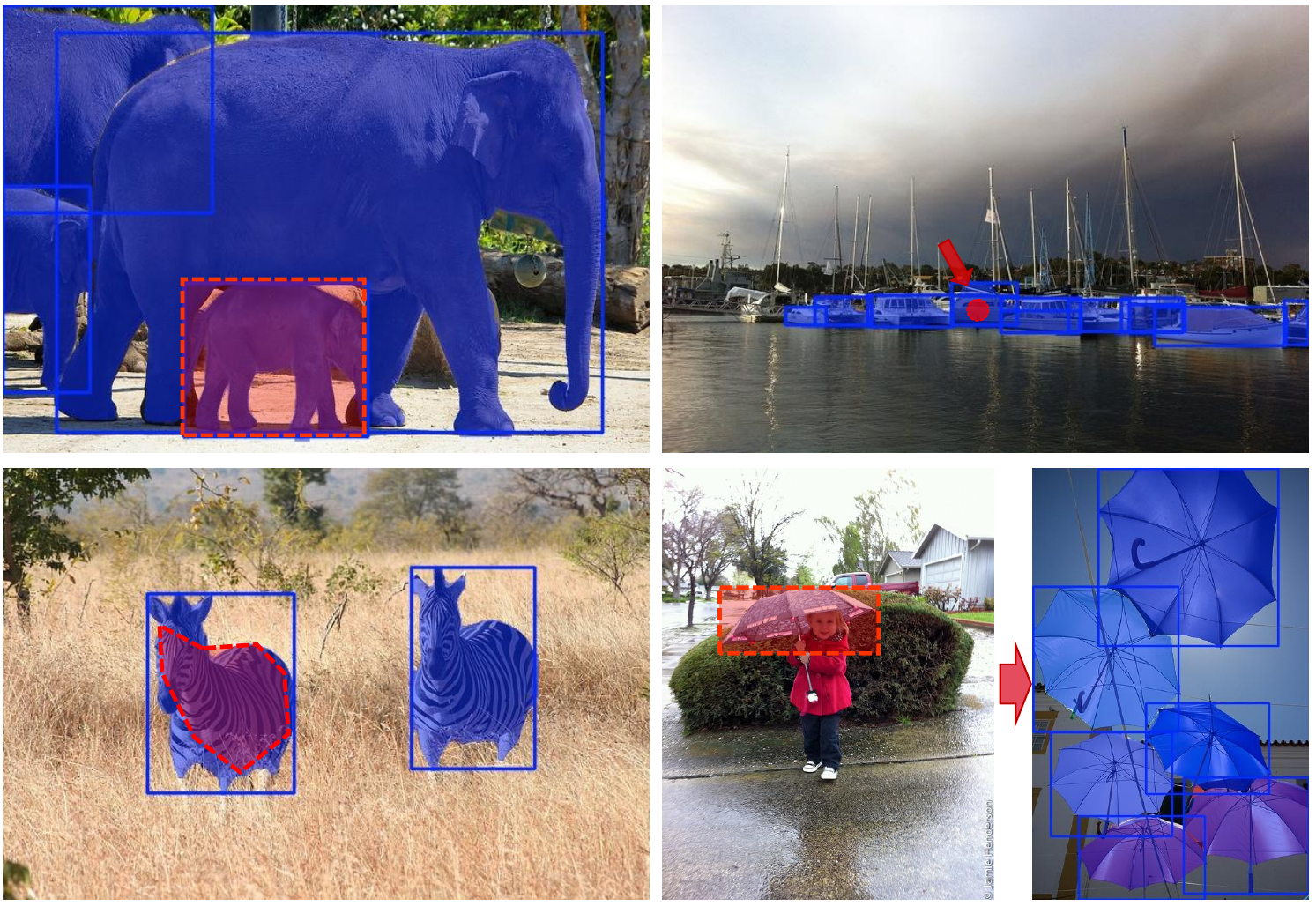}
    \caption{Prompt with visual inputs. YOLOE demonstrates the ability to identify objects guided by various visual prompts, like bounding box (top left), point (top right), handcrafted shape (bottom left). The visual prompt can also be applied across images (bottom right).}
    \label{fig:vp-visual}
    \vspace{-10pt}
\end{figure*}

\begin{figure*}[t]
    \centering
    \includegraphics[width=0.85\textwidth]{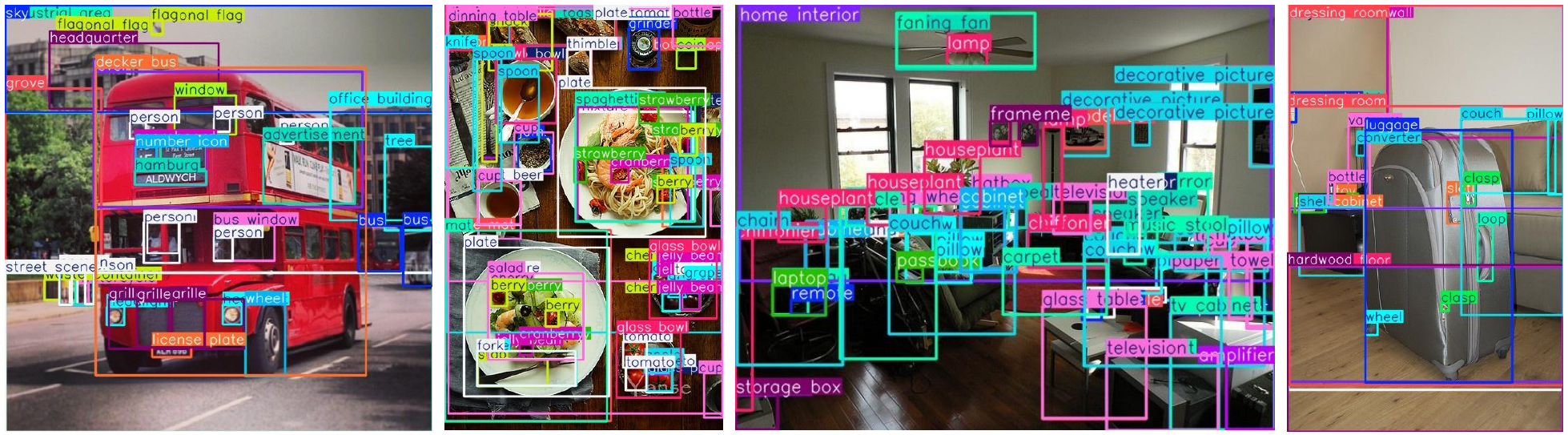}
    \caption{Prompt-free inference (omitting segmentation masks for clearer visualization). YOLOE can find diverse objects without prompt.}
    \label{fig:pf-visual}
    \vspace{-10pt}
\end{figure*}

\textbf{Prompt with visual inputs.} In \cref{fig:vp-visual}, we present the results of YOLOE with visual inputs as prompt. The visual inputs can take various forms, such as bounding box, point, or handcrafted shape. It can also be provided across the images. We can see that with visual prompt indicating the target object, YOLOE can accurately find other instances of the same category. Beside, it performs well across different objects and images, exhibiting robust capability.

\textbf{Prompt-free inference.} \cref{fig:pf-visual} shows the results of YOLOE with the prompt-free paradigm. We can see that in such setting, YOLOE achieves effective identification for diverse objects. This highlights its practicality in scenarios where predefined inputs are unavailable or impractical.

\end{document}